\PassOptionsToPackage{inline}{enumitem}

\documentclass[compactaffiliation]{Interspeech}

\interspeechcameraready

\usepackage{adjustbox}
\usepackage{booktabs}
\usepackage{tabularx}
\usepackage[inline]{enumitem}
\usepackage[utf8]{inputenc}
\usepackage{amsmath}
\usepackage{amssymb}
\usepackage{graphicx}
\usepackage{wrapfig}
\usepackage{xcolor}
\usepackage{multirow}
\usepackage{numprint}
\usepackage{nicefrac}
\usepackage{bbm}
\usepackage{ragged2e}
\usepackage{fdsymbol}
\usepackage{hyperref}
\usepackage[aboveskip=1pt,belowskip=1pt]{subcaption}
\usepackage{tipa}
\usepackage{phonetic}

\let\templatethebibliography=\thebibliography
\let\templateendthebibliography=\endthebibliography

\usepackage[numbers]{natbib}

\let\thebibliography=\templatethebibliography
\let\endthebibliography=\templateendthebibliography

\everypar{\looseness=-1}
\title{Phonetically-Augmented Discriminative Rescoring\\for Voice Search Error Correction}
\author[affiliation={1}]{Christophe}{Van Gysel}
\author[affiliation={1}]{Maggie}{Wu}
\author[affiliation={1}]{Lyan}{Verwimp}
\author[affiliation={1}]{Caglar}{Tirkaz}
\author[affiliation={1}]{\\Marco}{Bertola}
\author[affiliation={2, *}]{Zhihong}{Lei}
\author[affiliation={1}]{Youssef}{Oualil}

\affiliation{}{Apple}{USA}
\affiliation{}{Meta}{USA}

\email{\{cvangysel, maggiewu, lverwimp, c\_tirkaz, marco\_bertola, youalil\}@apple.com, leizh@meta.com}


\DeclareMathOperator*{\argmax}{arg\,max}
\newcolumntype{L}[1]{>{\hsize=#1\hsize\RaggedRight} X}

\begin{document}
\maketitle
{\renewcommand{\thefootnote}{* }\footnotetext{Work performed while at Apple.}}
\begin{abstract}
End-to-end (E2E) Automatic Speech Recognition (ASR) models are trained using paired audio-text samples that are expensive to obtain, since high-quality ground-truth data requires human annotators. Voice search applications, such as digital media players, leverage ASR to allow users to search by voice as opposed to an on-screen keyboard. However, recent or infrequent movie titles may not be sufficiently represented in the E2E ASR system's training data, and hence, may suffer poor recognition. %
In this paper, we propose a phonetic correction system that consists of (a) a phonetic search based on the ASR model's output that generates phonetic alternatives that may not be considered by the E2E system, and (b) a rescorer component that combines the ASR model recognition and the phonetic alternatives, and select a final system output. %
We find that our approach improves word error rate between 4.4 and 7.6\% relative on benchmarks of popular movie titles over a series of competitive baselines.
\end{abstract}

\newcommand{\PhoneticAM}{Monophone AM}
\newcommand{\AlternativeGenerationEngine}{PTT}

\section{Introduction}
\label{sec:introduction}

Modern Automatic Speech Recognition (ASR) systems consist of an end-to-end (E2E) neural model that outputs transcribed text directly given audio as input. E2E ASR models are trained using paired audio-text samples \citep{Prabhavalkar2023end} that are expensive to obtain. %

Digital media players allow users to playback digital content (e.g., movies) on a television, and to search by voice through a Virtual Assistant (VA) that uses ASR to transcribe a spoken query into text \citep{VanGysel2023modeling,Zhang2024entityrescoring,Saebi2021discriminative}.
Movie catalogs offered by streaming services are extensive and grow significantly over time. In 2024, $\sim$18k new movies were added to the Internet Movie Database (IMDb) \citep{IMDB2024Movies}. E2E ASR systems can have difficulty correctly recognizing entity names, such as movie titles, that are not well-represented in its paired training data---e.g., when a movie becomes popular after the training data cut-off date. %
Take as an example the query \emph{``play Pandorum''}, instructing the VA to play the 2009 Sci-fi/Horror movie \emph{Pandorum}, ranked within the top-5k most popular movies on IMDb in December 2024. Since the phrase \emph{Pandorum} is not common in English, and the process of obtaining new data and retraining the model is expensive, the ASR system misrecognizes the query as \emph{``play Pandora''}---instead instructing the VA to launch the \emph{Pandora} music app.

In addition to problems related to emerging entities, paired audio-text training data is often limited to the order of 100 million instances. %
While that may seem like an abundant amount of data, VAs handle multiple query types (e.g., device control, web search). An analysis of the randomly sampled, representative and anonymized user interactions of a popular VA deployed on digital media players in the U.S. shows that media catalog search makes up $\sim$12\% of traffic. %
Hence, under the assumption of 100M training samples, $\sim$12M of samples would contain media entity names and many queries will be repetitions of popular titles.
Since catalogs contain millions of titles, it is impossible to cover all titles in the training data.

In this paper, we present a method to correct the recognition result generated by an ASR system through a targeted phonetic search based on the ASR system's output. %
After the primary ASR model finishes decoding, a phonetic transcription is extracted and used as input for the search. %
After generating phonetic alternatives, we subsequently force align the alternatives to the audio using a phonetic acoustic model to determine their similarity to the audio. %
The N-best lists of the ASR system and the phonetic search are subsequently combined via a rescoring step that determines the final recognition.

Existing work on ASR correction consists of learning a token-to-token (T2T) translation model that generates correction candidates given the ASR model's inference output as input. The translation models take as input text hypotheses \citep{Guo2019spelling,Hrinchuk2020correction,Kang2024transformer} and optionally encoder features \citep{Hu2021transformer,Hu2022improving}, and they are trained on paired audio-text data obtained from annotators, possibly augmented with text-only data synthesized with Text-to-Speech (TTS)---requiring training sets between 40 and 200 million utterances \citep{Narayanan2019longformspeech,Hu2021transformer,Kang2024transformer}. %
While \citet{Guo2019spelling} point out that T2T approaches take into account the ASR system's characteristic error distribution, it introduces a tight coupling between ASR model and correction model that suffers from several shortcomings when deployed as part of an industrial application:
\begin{enumerate*}[label=(\arabic*)]
	\item (re-)training the correction model is expensive since it involves running ASR to generate a large amount of training data to learn a token-to-token mapping,
    \item reliance on TTS to inject text-only data (such as new entities) can bias the system towards errors that only occur in synthetic audio.
\end{enumerate*}
Consequently, data-hungry T2T correction approaches are impractical when deployed in an industrial application---since they can significantly increase the total wallclock time of system retraining and evaluation. %
An additional benefit of our approach is that it reuses many of the existing components that are part of the ASR system (\S\ref{sec:experiments:systems}) and provides accuracy gains without increasing asset size, which is important for on-device recognition \citep{Nussbaum2023application}.

To avoid the impracticalities of T2T correction models, our approach avoids a reliance on the ASR model's inference output to train its correction candidate component. Instead, we generate correction candidates using a phonetic lexicon search \citep{Logan2002documentretrieval,Saraclar2013empirical} that leverages the ASR system's existing external language model (LM) through a probabilistic interface---avoiding a tight coupling. At recognition time, the phonetic search runs after the primary E2E model finishes decoding and resembles a traditional hybrid ASR system that operates on phonetic units as input (as opposed to audio frames). %
The results from the phonetic search and the original ASR recognition are then combined using a discriminative rescorer \citep{Roark2004discriminative,Kobayashi2008discriminative} that is trained on a small set of paired audio-text data (in the order of $\sim$80k). %
The decomposition between correction generation, which is independent from the ASR system's output, and the rescorer that merges the correction alternatives with the ASR output, avoids the time-consuming and data-hungry nature of learning a token-to-token mapping from scratch.

\section{Methodology}
\label{sec:methodology}

\newcommand{\ASRNumAlternatives}{N}
\newcommand{\PTTNumAlternatives}{M}

Our approach operates within the decoding framework proposed by \citet{Lei2024amfusion}, where a subword-based Conformer CTC ASR model \citep{Graves2006ctc,Gulati2020conformer} is fused with a word-based external LM \citep{Laptev2022tlg} during the 1st pass, and a phoneme-based AM (referred to as "\PhoneticAM{}") is used during a 2nd pass to rescore an N-best list through forced alignment. The \PhoneticAM{} allows us to leverage external knowledge provided by a pronunciation lexicon.
As opposed to \citep{Lei2024amfusion}, which is limited to rescoring an N-best list generated by the 1st pass, we leverage the phonetic transcription generated by the \PhoneticAM{} to perform a narrow search for phonetically-similar hypotheses that are likely under the LM using our Phone-to-Text (\AlternativeGenerationEngine{}) component \citep{Logan2002documentretrieval,Saraclar2013empirical}. This allows us to generate hypotheses that were not considered by the 1st pass ASR model and that can leverage pronunciations from external sources (e.g., domain experts).
After obtaining phonetic alternatives, we combine the original N-best list obtained by the \PhoneticAM{} with the \AlternativeGenerationEngine{} alternatives and apply a multi-source rescorer model to select the best overall transcription.

\subsection{Phone-to-Text (\AlternativeGenerationEngine{})}
\label{sec:methods:ptt}

\newcommand{\ObservedPhoneticTranscription}{O}
\newcommand{\ObservedPhoneticUnit}{o}

\newcommand{\GroundTruthPhoneticTranscription}{\ObservedPhoneticTranscription{}^{*}}

\newcommand{\PhoneticUnits}{P}
\newcommand{\ExtendedPhoneticUnits}{\bar{\PhoneticUnits{}}}
\newcommand{\PhoneticUnit}{\MakeLowercase{\PhoneticUnits{}}}

\newcommand{\WordUnits}{V}

\newcommand{\Apply}[2]{#1\left(#2\right)}

\newcommand{\Prob}[1]{\Apply{P}{#1}}
\newcommand{\CondProb}[2]{\Prob{#1 \mid #2}}
\newcommand{\Length}[1]{\left|#1\right|}
\newcommand{\Exp}[1]{\Apply{\text{exp}}{#1}}

\newcommand{\InsertionProbability}{p_\text{ins}}

\begin{figure}
    \centering%
    \includegraphics[width=\columnwidth]{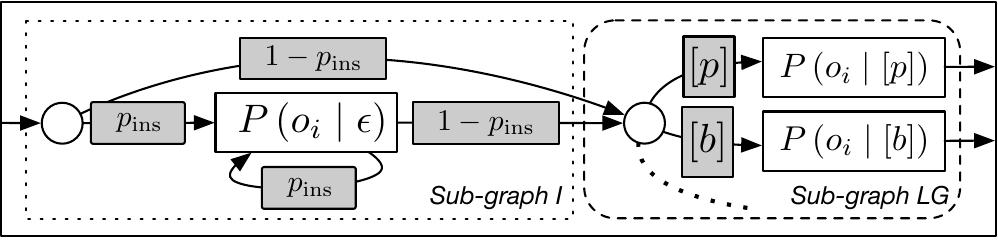}%
    \caption{Diagram representing a single step in our HMM. The dotted rectangle represents the component responsible for allowing symbols to occur in the observation, with probability $\InsertionProbability{}$, that are missing from the Phonetic LM. The dashed, rounded rectangle corresponds to a state in the Phonetic LM, with arcs labeled with phonetic units. The rectangular states are emitting states and are annotated with their emission probability distribution over $\ExtendedPhoneticUnits{} = \PhoneticUnits{} \cup \epsilon$. Each path in the HMM corresponds to a word sequence over vocabulary $\WordUnits{}$ that represents a phonetic alternative.\label{fig:ptt}}
\end{figure}

The component responsible for generating phonetic alternatives in our system is inspired by spelling correction systems in typed search \citep{Li2012spellingcorrection}, where a Hidden Markov Model (HMM) is used to model edits (i.e., insertions, deletions, substitutions \citep{Bahl1975noisychannel}) within the user-provided observed query with respect to a reference query set. In our work, we generate phonetic correction candidates by taking the phonetic transcription $\ObservedPhoneticTranscription{} = \ObservedPhoneticUnit{}_1, \ldots, \ObservedPhoneticUnit{}_n$ generated by the ASR as observation, and the hidden states correspond to phonetic LM transcriptions over a word-based vocabulary $\WordUnits{}$. %
The HMM is designed such that it supports edits from the phonetic representation of queries encoded by the phonetic LM, with respect to the ASR system's phonetic transcription $\ObservedPhoneticTranscription{}$.
Take as an example the query \emph{``Sherlock Holmes''} with IPA pronunciation \emph{``\textipa{"S 3 :r l \textturnscripta{} k "h o\rotOmega{} m z}''}, and the ASR system's recognized pronunciation \emph{``\textipa{"S 3 :r h a m z}''} corresponding to the transcription \emph{``Sherhams''}. In this case there would have been 3 deletions (\emph{``\textipa{l \textturnscripta{} k}''}) and 2 substitutions (\emph{\textipa{"h}} $\rightarrow$ \emph{\textipa{h}}, \emph{\textipa{o\rotOmega{}}} $\rightarrow$ \emph{\textipa{a}}).

\subsubsection{Hidden Markov Model}
\label{sec:methods:ptt:hmm}

Fig.~\ref{fig:ptt} depicts a single step in our HMM. %
The phonetic LM that makes up part of the hidden states of the HMM is obtained through the dynamic composition \citep{Allauzen2009generalized,Dixon2012dynamicwfst} of (a) a unigram lexicon Finite State Transducer (FST) \cite{Mohri2002fst}, $L$, that maps with phonetic units as input and words as output, and (b) a word-based Finite State Acceptor (FSA) \cite{Mohri2002fst}, $G$, that assigns a probability to a word sequence. 

The transition probabilities between states in $LG$ correspond to the LM probabilities, with the word-level probability mass in $L$ distributed across phonetic states through weight pushing \citep{Mohri2002fst}. %
After exploring an arc in the phonetic LM label with phonetic unit $\PhoneticUnit{} \in \PhoneticUnits{}$, we reach an emitting state with emission probability distribution $\CondProb{\ObservedPhoneticUnit{}}{\PhoneticUnit{}}$ (\S\ref{sec:methods:confusion}) where $\ObservedPhoneticUnit{} \in \ExtendedPhoneticUnits{}$ with $\ExtendedPhoneticUnits{} = \PhoneticUnits{} \cup \left\{\epsilon\right\}$. %
We allow to emit $\epsilon$ to support deletions, i.e., the observation is missing phone $\PhoneticUnit{}$ that occurs in the phonetic LM.
When $\ObservedPhoneticUnit{} \in \PhoneticUnits{}$, the emission probability represents the similarity between the observed phone $\ObservedPhoneticUnit{}$ and the phone $\PhoneticUnit{}$ that occurs within the LM, and models identity or substitution operations.

To support insertions, i.e., phones that occur in the observation and are unlikely under the phonetic LM, we prefix every state in $LG$ with a sub-graph $I$ that leads to an emitting state with probability $\InsertionProbability{}$ and with probability $1 - \InsertionProbability{}$ to its corresponding state in $LG$. The emitting state in $I$ has emission probability distribution $\CondProb{\ObservedPhoneticUnit{}}{\epsilon}$, allowing the HMM to accept observations without traversing arcs in the phonetic LM ($LG$).

\subsubsection{Phonetic confusion, deletion and insertion}
\label{sec:methods:confusion}

\newcommand{\DevelopmentSet}{U}
\newcommand{\DevelopmentInstance}{u}
\newcommand{\DevelopmentInstanceObs}{\ObservedPhoneticTranscription{}_\DevelopmentInstance{}}
\newcommand{\DevelopmentInstanceGT}{\ObservedPhoneticTranscription{}^{\prime}_\DevelopmentInstance{}}
\newcommand{\GroundTruthPhoneticUnit}{\ObservedPhoneticUnit^{\prime}}

\newcommand{\AlignFn}{A}

The emission probabilities $\CondProb{o}{o^\prime}$ with $o, o^\prime \in \ExtendedPhoneticUnits{}$ and insertion probability $\InsertionProbability{}$ are learned by aligning phonetic hypothesis and ground-truth transcriptions and estimating the confusion probabilities directly from the alignment \citep{Bahl1975noisychannel}.
For a set of development utterances $\DevelopmentInstance{} \in \DevelopmentSet{}$, we obtain phonetic transcription $\DevelopmentInstanceObs{}$ from our ASR system (\S\ref{sec:experiments:systems}) and align them to ground-truth phonetic transcription $\DevelopmentInstanceGT{}$ using Needleman-Wunsch (NW) \citep{Needleman1970general}, resulting in the multiset of alignment pairs $\Apply{\AlignFn{}}{\DevelopmentInstance{}} = \left\{ \left(\ObservedPhoneticUnit{}_i, \GroundTruthPhoneticUnit{}_i\right) \right\}_i$ between the two sequences with minimal edit distance. Symbols in $\DevelopmentInstanceObs{}$ that cannot be matched to any symbol in $\DevelopmentInstanceGT{}$, or vice versa, are aligned with $\epsilon$ (i.e., the gap symbol in the NW algorithm). %
We subsequently count the occurrence of alignment pairs across all utterances of $\DevelopmentSet{}$. Let $\Apply{\text{count}}{\ObservedPhoneticUnit{}, \GroundTruthPhoneticUnit{}}$ denote the occurrence of pair $\left( \ObservedPhoneticUnit{}, \GroundTruthPhoneticUnit{} \right)$ in multiset $\Apply{\AlignFn{}}{\DevelopmentSet{}} = \bigcup_{\DevelopmentInstance{} \in \DevelopmentSet{}} \Apply{\AlignFn{}}{\DevelopmentInstance{}}$, then we have:
\newcommand{\DevelopmentSetNumAlignmentPairs}{\Length{\Apply{\AlignFn{}}{\DevelopmentSet{}}}} %
{ %
\setlength{\abovedisplayskip}{3pt}
\setlength{\belowdisplayskip}{3pt}
\begin{equation*}
\CondProb{o}{o^\prime} = \frac{\Apply{\text{count}}{o, o^\prime}}{\sum_{p \in \ExtendedPhoneticUnits{}} \Apply{\text{count}}{p, o^\prime}},\quad\InsertionProbability{} = \frac{\sum_{p \in \ExtendedPhoneticUnits{}} \Apply{\text{count}}{p, \epsilon}}{\DevelopmentSetNumAlignmentPairs{}}
\end{equation*}%
}%
where $\DevelopmentSetNumAlignmentPairs{}$ is the number of alignment pairs across all utterances of $\DevelopmentSet{}$.

\subsubsection{Alternative generation}
\label{sec:methods:ptt:generation}

We generate an $\PTTNumAlternatives{}$-best list of phonetic alternatives after the primary ASR recognition process ends, by selecting the phonetic transcription from the ASR $\ASRNumAlternatives{}$-best with the largest likelihood under the \PhoneticAM{}, and subsequently running Viterbi decoding \citep{Viterbi1967error} on the HMM described above. %
The benefit of this approach is that the primary ASR process generates a phonetic transcription that is close to the audio, but may have an incorrect word-level transcription that causes recognition to return an incorrect result. By performing the additional targeted search based on phonetics, and using the lexicon/LM as driver (as opposed to the first pass E2E model), we can generate candidate recognitions with the correct word-level transcription. %
In the next section, we outline how we integrate the generated alternatives to improve the final ASR system recognition.

\subsection{Discriminative Rescoring}
\label{sec:methods:rescoring}

\newcommand{\CombinedNumAlternatives}{K}

\newcommand{\Hypothesis}{h}
\newcommand{\FeatureVector}{x}

\newcommand{\HypothesisASRFirst}{\Hypothesis{}_*}

\newcommand{\WeightVector}{w}

\newcommand{\RescoringScore}[1]{f\left(#1\right)}

\newcommand{\Utterance}{u}
\newcommand{\UtteranceGroundTruth}{\Utterance{}^*}

\newcommand{\MetricFn}{\Apply{\text{WER}}{\Hypothesis{}_i, \UtteranceGroundTruth{}}}

\newcommand{\CostFn}{C}
\newcommand{\Cost}[2]{\Apply{\CostFn{}_{\text{#1}}}{#2}}

\newcommand{\PhoneticCostFn}{\CostFn{}_{\text{Phon}}}
\newcommand{\AMCostFn}{\CostFn{}_{\text{Ac}}}

\newcommand{\PhoneticCost}[1]{\Apply{\PhoneticCostFn{}}{#1 \mid \HypothesisASRFirst{}}}
\newcommand{\AMCost}[1]{\Apply{\AMCostFn{}}{#1}}
\newcommand{\PhoneLength}[1]{\Length{\Apply{\text{Phone}}{#1}}}

\newcommand{\InterpolatedLmFn}{P_{\text{Interp}}}
\newcommand{\InterpolatedLmProb}[1]{\Apply{\InterpolatedLmFn{}}{#1}}
\newcommand{\LogInterpolatedLmProb}[1]{\log\InterpolatedLmProb{#1}}

\newcommand{\ComponentLmProbFn}{P}
\newcommand{\ComponentLm}{c_k}
\newcommand{\ComponentLmProb}[1]{\Apply{\ComponentLmProbFn{}}{#1 \mid \ComponentLm{}}}
\newcommand{\LogComponentLmProb}[1]{\log\ComponentLmProb{#1}}

\newcommand{\Indicator}[1]{\Apply{\mathbbm{1}}{#1}}

\newcommand{\ZScoreFn}{Z}
\newcommand{\ZScore}[1]{\Apply{\ZScoreFn}{#1}}

\newcommand{\GenericClipFn}{f}
\newcommand{\GenericClip}[1]{\Apply{\GenericClipFn{}}{0, #1}}

\newcommand{\MaxFn}{\text{max}}
\newcommand{\MinFn}{\text{min}}

\newcommand{\ClipAboveZero}[1]{\Apply{\MaxFn{}}{0, #1}}
\newcommand{\ClipBelowZero}[1]{\Apply{\MinFn{}}{0, #1}}

\begin{table}[t!]
    \caption{Overview of mathematical notation and features used during rescoring.} %
    \vspace*{-1\baselineskip}%
    \begin{subtable}[t]{\columnwidth}
    \caption{Legend of notation.}%
    \centering%
    \setlength\tabcolsep{3pt}%
    \footnotesize%
    \begin{tabularx}{\columnwidth}{l p{6cm}}%
    \toprule%
    \textbf{Notation} & \textbf{Description} \\
    \midrule%
    $\CombinedNumAlternatives{}$ & Combined number of top-hypotheses from the primary ASR process (top-$\ASRNumAlternatives{}$) and \AlternativeGenerationEngine{} (top-$\PTTNumAlternatives{}$) \\
    $\HypothesisASRFirst{}$ & Top-1 hypothesis of the ASR process \\
    \midrule%
    $\Indicator{\cdot}$ & Indicator function, evaluates to 1 if expression is true and 0 otherwise \\
    $\ZScore{\cdot}$ & Z-score normalization function applied over the hypotheses in the combined $\CombinedNumAlternatives{}$-best list \\
    \midrule%
    $\PhoneticCost{\Hypothesis{}}$ & Phonetic distance (\S\ref{sec:methods:confusion}) of hypothesis $\Hypothesis$ w.r.t. $\HypothesisASRFirst{}$ \\
    \midrule%
    $\PhoneLength{\Hypothesis{}}$ & Phone length of hypothesis $\Hypothesis{}$ provided by force-alignment with \PhoneticAM{} \\
    $\AMCost{\Hypothesis{}}$ & Acoustic cost of hypothesis $\Hypothesis{}$ through force-alignment with \PhoneticAM{} \\
    \midrule%
    $\ComponentLmProb{\Hypothesis{}}$ & LM probability of hypothesis $\Hypothesis{}$ under component $\ComponentLm{}$ \\
    $\InterpolatedLmProb{\Hypothesis{}}$ & Interpolated LM probability of hyp. $\Hypothesis{}$ using dynamic weights that maximize the prob. of the $\CombinedNumAlternatives{}$ hypotheses \\
    \bottomrule%
    \end{tabularx}
    \end{subtable} %
    \begin{subtable}[t]{\columnwidth} %
    \caption{Features used to represent hypotheses during discriminative rescoring. The numbers listed under "Transformations" refer to the rows of Table~\ref{tbl:transformations} and describe feature transformations applied on top of the base feature.\label{tbl:features}}%
    \centering%
    \setlength\tabcolsep{3pt}%
    \footnotesize%
    \begin{tabularx}{\columnwidth}{l L{1.0} l}%
    \toprule%
    \textbf{Category} & \textbf{Feature} & \textbf{Transformations} \\
    \midrule%
    Phonetic & $\PhoneticCost{\Hypothesis{}_i}$ & 1 \\
             & $\PhoneLength{\Hypothesis{}}$ & 2 \\
    Acoustic & $\AMCost{\Hypothesis{}_i}$ & 2, 3, 4 \\
    Interp. LM & $\LogInterpolatedLmProb{\Hypothesis{}_i}$ & 2, 3, 4 \\
    Comp. LM & $\LogComponentLmProb{\Hypothesis{}_i}$ & 4, 5 with $t = \Apply{\log}{10^{-7}}$ \\
    Other & $\Indicator{\Apply{\text{source}}{\Hypothesis{}_i} = x}$ with $x \in \{\text{ASR}, \text{PTT}\}$ & - \\
    \bottomrule%
    \end{tabularx} %
    \end{subtable} %
\end{table}

\begin{table*}[t!]
    \newcommand{\FeatureFn}{f}%
    \newcommand{\Feature}[1]{\Apply{\FeatureFn{}}{#1}}%
    \caption{Transformations applied on top of features. Function $\Feature{\Hypothesis{}}$ in this table represents any of the features described in Table~\ref{tbl:features}, where any aggregation operation iterates over the combined $\CombinedNumAlternatives{}$ hypotheses to be rescored.\label{tbl:transformations}}%
    \centering%
    \setlength\tabcolsep{3pt}%
    \footnotesize%
    \begin{tabularx}{\textwidth}{l l l p{9cm}}%
    \toprule%
    & \textbf{Transformation} & \textbf{Type} & \textbf{Description} \\
    \midrule%
    1 & $\Indicator{\Feature{\Hypothesis{}_i} = \text{min}_j\left\{\Feature{\Hypothesis{}_j}\right\}}$ & Binary & Feature value $\Feature{\Hypothesis{}_i}$ is smallest value among all hypotheses \\
    2 & $\Apply{g}{0, \Feature{\Hypothesis{}_i} - \Feature{\HypothesisASRFirst{}}}$ with $g \in \{\text{min}, \text{max}\}$ & Continuous & The difference between feature value $\Feature{\Hypothesis{}_i}$ and feature value $\Feature{\HypothesisASRFirst{}}$ of the top-1 ASR hyp., clipped above/below zero \\
    3 & $\Indicator{\Feature{\Hypothesis{}_i} \oplus \Feature{\HypothesisASRFirst{}}}$ with $\oplus \in \left\{=, <, >\right\}$ & Binary & Feature value $\Feature{\Hypothesis{}_i}$ is larger, less, or equal to the feature value $\Feature{\HypothesisASRFirst{}}$ of the top-1 ASR hyp. \\
    4 & $\Apply{g}{0, \ZScore{\Feature{\Hypothesis{}_i}}}$ with $g \in \{\text{min}, \text{max}\}$ & Continuous & Standardized feature value $\Apply{Z}{\Feature{\Hypothesis{}_i}}$ clipped above/below zero \\
    5 & $\Indicator{\Apply{\MaxFn{}_j}{\Feature{\Hypothesis{}_j}} \oplus t}$ with $\oplus \in \{<, >\}$ & Binary & The largest feature value among all hypotheses is larger/less than a predetermined threshold $t$ \\
    \bottomrule%
    \end{tabularx}
\end{table*}

After obtaining the top-\PTTNumAlternatives{} alternatives from \AlternativeGenerationEngine{}, we combine the obtained \PTTNumAlternatives{} alternatives with the the top-\ASRNumAlternatives{} alternatives from the primary speech recognition process into a combined set of $\CombinedNumAlternatives{} = \ASRNumAlternatives{} + \PTTNumAlternatives{}$ alternatives for utterance $\Utterance{}$, and select a final recognition through discriminative rescoring \citep{Roark2004discriminative}. %
More specifically, we represent each hypothesis $\Hypothesis{}_i$ by a feature vector $\FeatureVector{}_i$ consisting of features extracted from the \PhoneticAM{} force-alignment process, the LMs, and more; see Table~\ref{tbl:features} for an overview. %
Feature values are standardized using statistics computed on the training set. In addition, we include higher-order composite features obtained by multiplying individual features to allow the model to capture non-linear relationships.

For each hypothesis $\Hypothesis{}_i$, we compute its score $\RescoringScore{\Hypothesis{}_i} = \FeatureVector{}_i \cdot \WeightVector{}^\intercal$ through a weighted combination with weight vector $\WeightVector{}$ that is optimized through a minimium WER objective \citep{Prabhavalkar2018minwer}.
Given utterance $\Utterance{}$ and its ground truth transcription $\UtteranceGroundTruth{}$, its contribution to the loss function equals: %
{ %
\setlength{\abovedisplayskip}{3pt}
\setlength{\belowdisplayskip}{3pt}
\begin{equation}
\label{eq:instance_loss}
l\left( \Utterance{}, \UtteranceGroundTruth{} \right) = \sum^{\CombinedNumAlternatives{}}_{i = 1} \CondProb{\Hypothesis{}_i}{\Utterance} \cdot \MetricFn{}
\end{equation} %
} %
where $\CondProb{\Hypothesis{}_i}{\Utterance} = \nicefrac{\Exp{\RescoringScore{\Hypothesis{}_i}}}{\sum^{\CombinedNumAlternatives{}}_{j = 1} \Exp{\RescoringScore{\Hypothesis{}_j}}}$ denotes a distribution over hypotheses for utterance $\Utterance{}$, and $\MetricFn{}$ is the normalized word-error rate (WER) between hypothesis $\Hypothesis{}_i$ and ground-truth $\UtteranceGroundTruth{}$ clipped within $(0, 1)$. Eq.~\ref{eq:instance_loss} is minimized using batched gradient descent with the Adam optimizer \citep{Kingma2014Adam}. Utterances for which all $\Hypothesis{}_i$ have the same WER are discarded during training. At recognition time, the hypothesis with the highest score, i.e., $\argmax_{\Hypothesis{}_i} \RescoringScore{\Hypothesis{}_i}$, is chosen as the final output.

\section{Experimental set-up}

\newcommand{\Identifier}[1]{{\footnotesize\texttt{#1}}}

\newcommand{\SystemBase}{\Identifier{base}}
\newcommand{\SystemLmRescoring}{\Identifier{lmrescoring}}
\newcommand{\SystemAmFusion}{\Identifier{amfusion}}
\newcommand{\SystemCorrections}{\Identifier{corrections}}

\newcommand{\PlusSystem}{{\Large$\Rdsh$} $+\text{ }$} %

\subsection{Systems under comparison}
\label{sec:experiments:systems}

\begin{table}[t!]
    \newcolumntype{R}[2]{%
        >{\adjustbox{angle=#1,lap=\width-(#2)}\bgroup}%
        l%
        <{\egroup}%
    }%
    \newcommand*\rot{\multicolumn{1}{R{35}{0.5em}}}%
    \caption{Overview of compared systems and the models/approaches they leverage. %
    \label{tbl:system_comparison}}%
    \centering%
    \footnotesize%
    \renewcommand{\arraystretch}{0.5}%
    \setlength{\tabcolsep}{8pt}%
    \resizebox{0.95\columnwidth}{!}{%
    \begin{tabularx}{\columnwidth}{lccccc}%
    \toprule%
    \textbf{System} & \rot{\scriptsize Conformer/TLG} & \rot{\scriptsize External LM} & \rot{\scriptsize Monophone AM} & \rot{\scriptsize PTT (\S\ref{sec:methods:ptt})} & \rot{\scriptsize Rescorer (\S\ref{sec:methods:rescoring})} \\
    \midrule%
    \SystemBase{} & $\checkmark$ & $\checkmark$ & & & \\
    \PlusSystem{}\SystemLmRescoring{} & $\checkmark$ & $\checkmark$ & & & \\
    \PlusSystem{}\SystemAmFusion{} & $\checkmark$ & $\checkmark$ & $\checkmark$ & & \\
    \PlusSystem{}\SystemCorrections{} & $\checkmark$ & $\checkmark$ & $\checkmark$ & $\checkmark$ & $\checkmark$ \\
    \bottomrule%
    \end{tabularx}}
\end{table}

We compare the following speech recognition systems where each system builds upon the previous one---each adding a component that aims to improve spoken entity query accuracy (see Table~\ref{tbl:system_comparison} for an overview): %
\begin{enumerate*}[label=(\arabic*)]
    \item Our \SystemBase{} system, a sub-word ($8$k units) Conformer CTC model \citep{Graves2006ctc} with a TLG graph decoder (using $T_\text{compact}$ \citep{Laptev2022tlg}, a lexicon of $600$k words, internal LM subtraction \citep{Meng2021ilm}, and a beam width of $12$) that linearly interpolates (a) a FOFE NNLM \cite{Zhang2015fixed,Nussbaum2023application} trained on a mixture of usage-based and synthetic VA queries with a truncated context window of 3, and (b) an N-gram LM trained on synthetic entity-centric VA queries \citep{VanGysel2022phirtn,Gondala2021error,Pusateri2019connecting} using weights $\left(0.9, 0.1\right)$.
    \item The previous system + \SystemLmRescoring{} on top, which rescores the lattice with the same LMs used during the 1st pass, but where the NNLM uses a context window of 7, and the interpolation weights are determined dynamically by maximizing the likelihood of the top-5 hypotheses in the lattice.
    \item The previous system + \SystemAmFusion{} \citep{Lei2024amfusion} on top to rescore the N-best ($N = 25$) by force-aligning the Conformer from the 1st pass and a monophone AM trained using cross-entropy.
    \item The previous system + our method, the \SystemCorrections{} alternative generation and rescoring components proposed in this paper, where \AlternativeGenerationEngine{} uses the n-gram LM from (1) as graph $G$ (\S\ref{sec:methods:ptt:hmm}) with a beam width of $15$ (\S\ref{sec:methods:ptt:generation}); and the rescoring engine leverages the NNLM/N-gram from (2), and the monophone AM from (3); we rescore $\CombinedNumAlternatives{} = 11$ hypotheses ($\ASRNumAlternatives{} = 1, \PTTNumAlternatives{} = 10$) (\S\ref{sec:methods:rescoring}).
\end{enumerate*}

\subsection{Evaluation}
\label{sec:experiments:evaluation}

\newcommand{\GeneralVA}{\Identifier{GeneralVA}}
\newcommand{\MovieVoiceSearchPlay}{\Identifier{PlayMovie}}
\newcommand{\MovieVoiceSearchVerbless}{\Identifier{VerblessMovie}}

We evaluate the systems mentioned in \S\ref{sec:experiments:systems} on two categories of English VA queries aimed at a digital media player focused on TV shows and movies in the U.S./Canada: %
\begin{enumerate*}[label=(\arabic*)]
    \item \GeneralVA{}: 41k human-graded, randomly sampled, representative and anonymized historical user utterances issued at the VA of a popular digital media player that spans a period of more than 1 year and with a cut-off date in March 2024 covering several use-cases (e.g., device control, web/media search),
    \item human-recorded queries instructing the digital media player to play a movie, according to the top-K most popular movie titles on a popular streaming service in April 2024, with two variations: %
    \begin{enumerate*}[label=(\alph*)]
        \item \MovieVoiceSearchPlay{}: 2k titles with queries of the form \emph{``play \$TITLE''}, representing the use-case of the user issuing the query from any screen in the system, and
        \item \MovieVoiceSearchVerbless{}: 4k titles where the query is only the movie title, representing the use-case of the user dictating the title into a movie search bar.
    \end{enumerate*}
\end{enumerate*}

\subsection{Training data}
\label{sec:experiments:training}

\noindent%
\textbf{Conformer and monophone acoustic models. } %
The Conformer/monophone AMs are both trained on the same collection of over 10M hours of annotator transcribed audio obtained from randomly sampled, representative and anonymized spoken VA queries with a cut-off date in March 2024. The utterances are sampled from the same population as our evaluation set (\S\ref{sec:experiments:evaluation}), but without speaker overlap.

\noindent%
\textbf{Language model.} %
The NNLM and N-gram are trained on query logs from a randomly sampled, representative and anonymized spoken VA queries, augmented with entity-centric synthetic queries \citep{VanGysel2022phirtn,Sannigrahi2024synthetic} spanning several VA domains (e.g., music, TV).

\noindent%
\textbf{Rescorer and phonetic confusion.} %
The rescoring component introduced in \S\ref{sec:methods:rescoring} is trained on a mixture of %
\begin{enumerate*}[label=(\arabic*)]
    \item 40k human-graded digital media player VA queries sampled from the same population as our test set queries (\S\ref{sec:experiments:evaluation}), and
    \item 40k human-recorded synthetic entity-centric queries from 2020 and prior, spanning several domains (music, movies).
\end{enumerate*}

\section{Results}

\begin{table}[t!]
    \caption{Overview of our experimental results where each system adds a component upon the previous one (\S\ref{sec:experiments:systems}).\label{tbl:results}}%
    \centering%
    \renewcommand{\arraystretch}{0.5}%
    \setlength{\tabcolsep}{3pt}%
    \resizebox{0.95\columnwidth}{!}{%
\begin{tabular}{@{}llll@{}}%
\toprule%
\textbf{System}&\GeneralVA{}&\MovieVoiceSearchPlay{}&\MovieVoiceSearchVerbless{}\\%
\midrule%
\SystemBase{}&8.74 \phantom{\small $\left(0.00\%\right)$}&5.72 \phantom{\small $\left(0.00\%\right)$}&6.96 \phantom{\small $\left(0.00\%\right)$}\\%
\midrule%
\PlusSystem{}\SystemLmRescoring{}&8.54 \phantom{\small $\left(0.00\%\right)$}&5.39 \phantom{\small $\left(0.00\%\right)$}&6.52 \phantom{\small $\left(0.00\%\right)$}\\%
\PlusSystem{}\SystemAmFusion{}&5.92 \phantom{\small $\left(0.00\%\right)$}&4.79 \phantom{\small $\left(0.00\%\right)$}&5.80 \phantom{\small $\left(0.00\%\right)$}\\%
\midrule%
\PlusSystem{}\SystemCorrections{}&5.92 \phantom{\small $\left(0.00\%\right)$}&4.58 {\small $\left(4.40\%\right)$}&5.37 {\small $\left(7.55\%\right)$}\\\bottomrule%
\end{tabular}%
    }%
\end{table}

Table~\ref{tbl:results} shows the results of our experiments, where each row represents a system that includes the components in the rows that precede it. %
The \SystemBase{} system provides a baseline single-pass Conformer-driven graph decoding set-up (\S\ref{sec:experiments:systems}). Second pass methods, which rescore the results of the first pass, improve upon the result by taking into account additional context (\SystemLmRescoring{}) or additional signals (\SystemAmFusion{}). However, a major drawback of second pass methods that only perform a rescoring is that they only operate on the top hypotheses generated by the first pass, and are unable to resolve recognition errors that occur due to sparsity in the paired audio-text training data (\S\ref{sec:introduction}).
Our method (last row of Table~\ref{tbl:results}) improves over the system that combines the 1st pass, LM rescoring and AM fusion (\S\ref{sec:experiments:systems}) on movie voice search queries between $4.4\%$ (\MovieVoiceSearchPlay{}) and $7.55\%$ (\MovieVoiceSearchVerbless{}) relative WER. As expected, correction does not negatively impact WER on \GeneralVA{}---which does not contain recent entities (\S\ref{sec:experiments:evaluation}). %
The oracle WERs of the phonetically-augmented $\CombinedNumAlternatives{}$-best lists, used within our corrections approach, are $4.0$, $2.94$ and $3.47$ for the three test sets in Table~\ref{tbl:results} respectively.

\begin{figure}
    \centering%
    \includegraphics[width=\columnwidth]{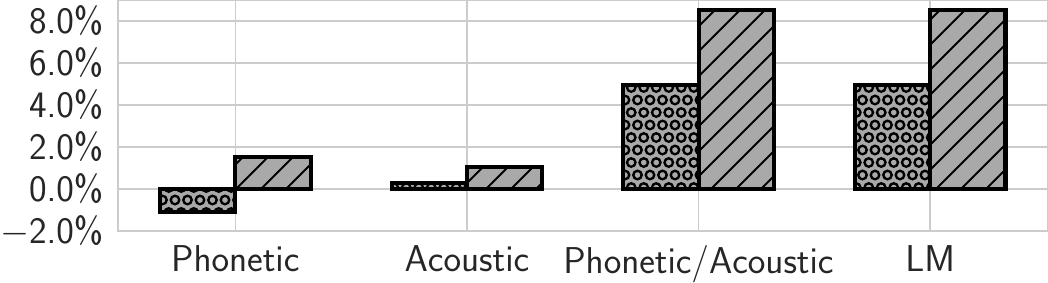}%
    \caption{Ablation study demonstrating the rel. WER increase that occurs when removing primary feature groups (see Table~\ref{tbl:features}) on the \MovieVoiceSearchPlay{} (circles) and \MovieVoiceSearchVerbless{} (stripes) test sets.\label{fig:ablation}}
\end{figure}

Fig.~\ref{fig:ablation} shows the results of an ablation study where we removed primary feature groups during training of the rescoring model, across the \MovieVoiceSearchVerbless{} and \MovieVoiceSearchPlay{} movie search test sets that each represent a different use-case (\S\ref{sec:experiments:evaluation}). %
We see that features extracted from the LM have the largest impact when removed during training, and most of the gains provided by our approach vanish when LM features are removed. In addition, when we remove both phonetic and acoustic features, performance likewise degrades significantly. %
When considering the phonetic and acoustic feature groups separately, we see differing behavior based on test set use-case. For the \MovieVoiceSearchVerbless{} use-case, removing phonetic features leads to a degradation. For the use-case where movie titles are prefixed with left-context \emph{``play''} (\MovieVoiceSearchPlay{}), the roles are reversed and removal of the phonetic features improves quality slightly. %
Our finding is that both the LM and acoustic/phonetic features are necessary. Without acoustic/phonetic signals, it is impossible to measure similarity of a correction hypothesis to the actual audio, and hence, alternatives with a high likelihood under the LM would always be chosen. Inversely, without LM features, it is difficult to select the most likely candidate amongst several hypotheses that are all phonetically/acousticly close.

\section{Conclusions}

We introduced a phonetic correction system that generates phonetic alternatives (PTT) based on the ASR system's recognition, followed by a discriminative rescorer that selects the best alternative amongst the ASR and PPT N-best lists. %
PTT allows to generate correction candidates without the need for large-scale audio-text samples, and avoids a tight coupling with the ASR system (as opposed to token-to-token models trained on ASR errors). The feature-based rescorer component, trained on a small amount of audio-text data, then subsequently combines the top hypotheses from ASR and PTT, and selects a final recognition. The two-staged design is practical in industrial ASR deployments (\S\ref{sec:introduction}). An additional benefit is that the our approach reuses many components already present in the ASR system (Table~\ref{tbl:system_comparison}) and does not lead to a significant asset size increase.
Future work includes improved phonetic alternative methods, that similarly do not create a tight coupling with the primary ASR model, and enhanced rescoring models/features.

\section{Acknowledgments}

We thank %
Leo Liu, %
Linda Arsenault, %
Man-Hung Siu, %
Sameer Badaskar, %
Stephen Pulman, %
Takaaki Hori, %
Thiago Fraga da Silva, %
and the anonymous reviewers %
for their comments and feedback.

\bibliographystyle{IEEEtranN}
\bibliography{interspeech2025-voice_search_correction}

\end{document}